\documentclass[conference]{IEEEtran}
\IEEEoverridecommandlockouts
\usepackage{cite}
\usepackage{amsmath,amssymb,amsfonts}
\usepackage{algorithmic}
\usepackage{graphicx}
\usepackage{textcomp}
\usepackage{xcolor}
\usepackage{multirow}
\usepackage{algorithmic}
\usepackage{algorithm}
\usepackage{subcaption}
\usepackage{booktabs}
\usepackage{hyperref}
\def\BibTeX{{\rm B\kern-.05em{\sc i\kern-.025em b}\kern-.08em
    T\kern-.1667em\lower.7ex\hbox{E}\kern-.125emX}}
\begin{document}

\title{Enhancing Diffusion-based Dataset Distillation via Adversary-Guided Curriculum Sampling
\thanks{$^*$Corresponding author

This study is supported by the National Natural Science Foundation of China (Grant No. 62306084, U23B2051), and Shenzhen Science and Technology Program (Grant No. GXWD20231128102243003, KJZD20230923115113026, and ZDSYS20230626091203008), and China Postdoctoral Science Foundation (Grant No. 2024M764192).}
}

\author{
Lexiao Zou\textsuperscript{1} \quad
Gongwei Chen\textsuperscript{1} \quad
Yanda Chen\textsuperscript{1} \quad
Miao Zhang\textsuperscript{1}$^*$ \vspace{.2em} \\
\textsuperscript{1}Harbin Institute of Technology, Shenzhen\vspace{.2em} \\
Email: \{lokshawchau, cydaaa30\}@gmail.com \quad \{chengongwei, zhangmiao\}@hit.edu.cn
}

\maketitle

\begin{abstract}
Dataset distillation aims to encapsulate the rich information contained in dataset into a compact distilled dataset but it faces performance degradation as the image-per-class (IPC) setting or image resolution grows larger. Recent advancements demonstrate that integrating diffusion generative models can effectively facilitate the compression of large-scale datasets while maintaining efficiency due to their superiority in matching data distribution and summarizing representative patterns. However, images sampled from diffusion models are always blamed for lack of diversity which may lead to information redundancy when multiple independent sampled images are aggregated as a distilled dataset. To address this issue, we propose Adversary-guided Curriculum Sampling (ACS), which partitions the distilled dataset into multiple curricula. For generating each curriculum, ACS guides diffusion sampling process by an adversarial loss to challenge a discriminator trained on sampled images, thus mitigating information overlap between curricula and fostering a more diverse distilled dataset. Additionally, as the discriminator evolves with the progression of curricula, ACS generates images from simpler to more complex, ensuring efficient and systematic coverage of target data informational spectrum. Extensive experiments demonstrate the effectiveness of ACS, which achieves substantial improvements of 4.1\% on Imagewoof and 2.1\% on ImageNet-1k over the state-of-the-art.
\end{abstract}


\section{INTRODUCTION}
\label{sec:intro}
% background
As the volume of multimedia data continues to proliferate, it provides expansive corpus for multimedia research. However, this abundance also imposes significant challenges on storage and computational resources\cite{DD,ImageNet,DiT}. Dataset distillation is proposed to compress the information contained within large-scale datasets into a small but more compact synthetic dataset, thereby maintaining test performance when trained on the distilled dataset compared to the original dataset\cite{DD}.

% related work - bi-level optimization
Early approaches frame dataset distillation as a bi-level optimization task. They train a model with the synthetic dataset in the inner loop while optimizes synthetic pixels in the outer loop by designed loss functions based on the performance of the inner-loop model\cite{DD}. These methodologies encompass techniques such as back propagation through time\cite{RaT-BPTT, RTP}, gradient matching\cite{gm, cafe, IDC}, distribution matching\cite{DM}, training trajectory matching\cite{Cazenavette2022DMTT,Cui2022TESLA,Du2022FTD,DATM}. However, due to their reliance on pixel-wise parameterization and the bi-level optimization process, the number of parameters and the computational complexity of optimization increase significantly with the number and resolution of images of the synthetic dataset. 
%For example, DATM\cite{DATM} requires 9.6 hours of computation time on four 80GB NVIDIA A100 GPUs to distill a subset of 1000 images per class (IPC) from the CIFAR-10 dataset, each at a resolution of 32x32.%
Consequently, the applicability of these approaches is largely confined to small-scale, low-resolution datasets like CIFAR\cite{CIFAR} and Tiny-Imagenet\cite{ImageNet}. 

% diffusion-based distillation
To achieve the scalability of dataset distillation, Minimax\cite{Gu2023minimax} uncovers the natural compression capabilities of synthetic datasets composed of images sampled from diffusion models\cite{DiT} trained on target dataset. Training the diffusion model implicitly optimizes an upper bound of the distribution discrepancy between synthetic and real data, which is proven a potent objective for dataset distillation\cite{Yuan2023RealFake, DM}. Building upon this foundation, Minimax introduces minimax criteria for diffusion training to enhance the diversity and representativeness of the distribution modeled by the diffusion process. This approach requires only fine-tuning a pre-trained diffusion model on the target dataset and then constructing the synthetic dataset by uniformly sampling images from the fine-tuned diffusion model. In summary, diffusion-based dataset distillation significantly reduces computational requirements, facilitating further research in complex and high-resolution datasets like ImageNet\cite{ImageNet}. However, Minimax overlooks the ``low temperature" nature of diffusion sampling, i.e., focusing the generation on well-learned high-probability region\cite{diffdiver,genlowdensity}. For dataset distillation, the low-temperature nature becomes particularly pronounced since individual sampling is carried multiple times to constitute a distilled dataset. Similar typical patterns in such distilled dataset lead to suboptimal utilization of distillation budget. 
\begin{figure*}[h]
\centering
\includegraphics[width=15.25cm]{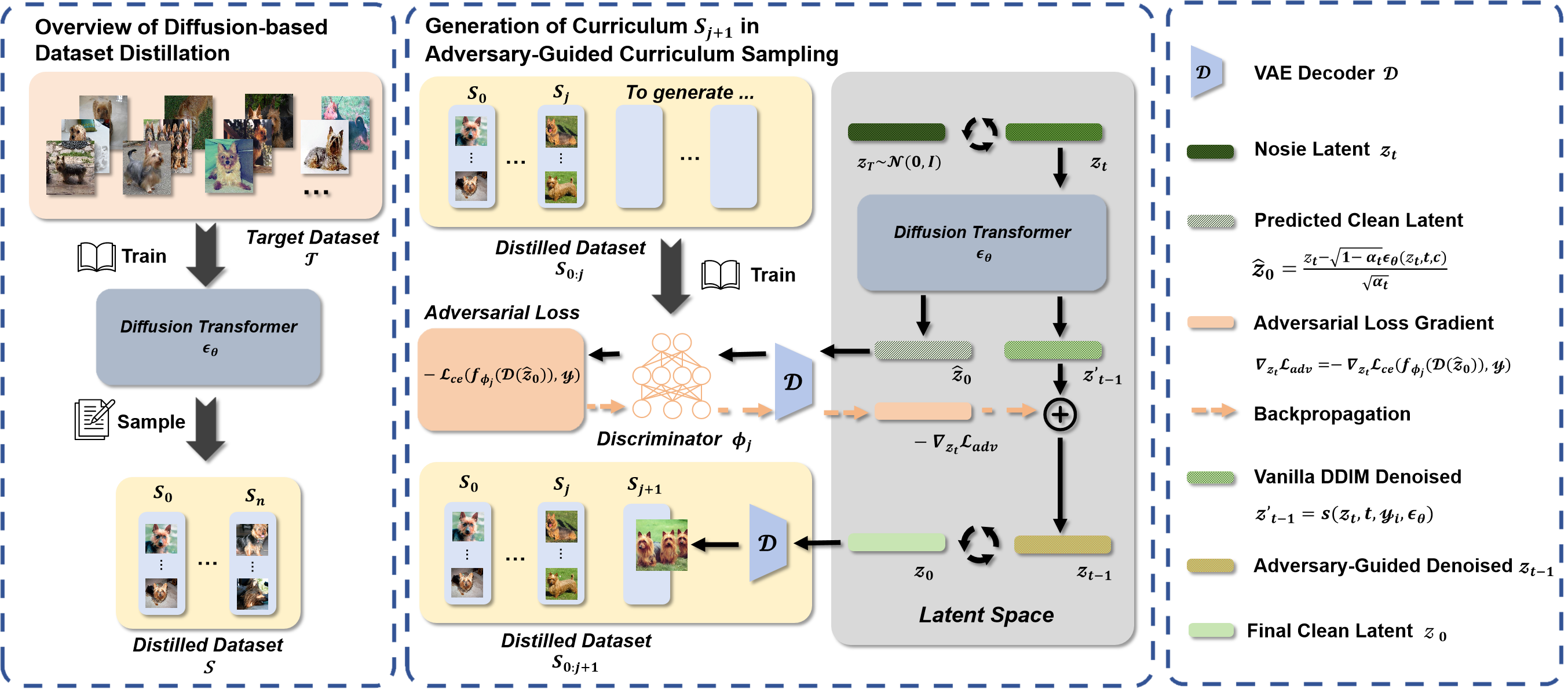}
\caption{The illustration of our proposed Adversary-guided Curriculum Sampling. ACS is applied in the sampling stage of diffusion-based dataset distillation. It divides the sampling of distilled dataset $S$ into $n$ curricula $S=S_{0:n}=\bigcup_{i=0}^{n} S_i$. For sampling an image in curriculum $S_{j+1}$, ACS first trains a discriminator $\phi_j$ with all preceding curricula $S_{0:j}$. Then the adversarial loss guides diffusion denoising steps towards generating sample that challenges discriminator $\phi_j$.}
\label{fig:architecture}
\end{figure*}
To tackle the aforementioned limitations, we introduce Adversary-guided Curriculum Sampling (ACS), a curriculum diffusion sampling framework ensuring that the synthetic dataset not only grows in size but also becomes increasingly representative of target data diversity. Our approach divides distilled dataset into multiple sequential curricula. For generating each curriculum, we first leverage the synthetic data generated from all preceding curricula to train a discriminator. Then we employ an adversarial loss to guide the diffusion sampling process toward producing images that challenge the discriminator. By progressively sampling in this manner, our method seeks to reduce information redundancy between curricula thus promises efficient coverage of target data distribution.

Notably, ACS generates samples from simple to complex as curriculum advances due to the evolvement of discriminator, which ensures systematically recover target dataset information spectrum. Recent researches\cite{Selmatch, DATM,CurriculumLearning, cudd, chen2025influenceguided, chen2025ccfs} highlight that as the number of images per class increases, the synthetic dataset should progressively encompass more complex and diverse features of the real dataset, maintaining an appropriate level of difficulty. In our work, we observe that in the initial curriculum, the small size of the synthetic dataset limits the performance of the discriminator trained on it. Consequently, the diffusion deoising function plays a dominant role in the sampling process, favoring the generation of samples in well-learned high-probability region from target data manifold\cite{Gu2023minimax, genlowdensity}. As the volume of distilled data grows, the performance of the discriminator steadily improves which promotes the gradual emergence of less common and more challenging samples. 

Extensive experiments on ImageNet-1k\cite{ImageNet} and its challenging subset demonstrate that our approach consistently enhances diffusion-based dataset distillation methods across different image-per-class setting and downstream test model architecture. Specifically, ACS surpasses current state-of-the-art method by 4.1\% on Imagewoof and 2.1\% on ImageNet-1k. Analysis on distilled images complexity and feature distribution further demonstrate that our method gains a diverse distilled dataset by sampling images from simple to complex across curriculum.

\section{METHODS}

\subsection{Diffusion for Dataset Distillation}

Dataset distillation\cite{DD} aims to distill a small surrogate dataset $\mathcal{S}=\left\{\left(x_i, y_i\right)\right\}_{i=1}^{N_S}$ from large scale target dataset  $\mathcal{\mathcal { T }}=\left\{\left(\mathbf{x}_i, y_i\right)\right\}_{i=1}^{N_T}$, where $N_s\ll N_T$,  that classification model trained on $\mathcal{S}$ achieves comparable performance with that on $\mathcal{T}$. Here,  $x_i$ and $y_i$ denote an image and its corresponding class label. Unlike traditional dataset distillation methods, which treat all data points $x_i$ in the distilled dataset $\mathcal{S}$ as parameters and optimize them with bi-level optimization, diffusion-based dataset distillation\cite{Gu2023minimax} adopts a fundamentally different approach. Specifically, we first train a class-conditioned diffusion model on the target dataset to capture intricate patterns and distributions. Once trained, diffusion sampling conditioned on each class $y_i$ from the diffusion model is performed $N_S$ times to construct the distilled dataset $S$. 

%Diffusion models\cite{DDIM} learn target data distributions by gradually adding Gaussian noise and subsequently reversing this process. This mechanism effectively squeeze the information from the target data into the diffusion model, enabling it to capture intricate patterns and distributions. 
In this work, we adopt DiT\cite{DiT} as the backbone. An encoder $E$ maps images $x$ to latent codes $z$ while a decoder $D$ reconstructs these codes back to the image domain $\tilde{x}=D(z)$. Diffusion forward noising process introduces Gaussian noise $\epsilon \sim \mathcal{N}(0,\mathrm{I})$ to original latent code $z_0$ over time step $t$: $z_t=\sqrt{\alpha_t} z_0+\sqrt{1-\alpha_t} \epsilon$, where $\alpha_t$ is a hyper-parameter. Training minimizes the squared error between predicted noise $\epsilon_\theta\left(z_t,t,c\right)$ and the ground truth $\epsilon$.  For sampling each image, we follow Denoising Diffusion Implicit Model (DDIM)\cite{DDIM}. DDIM iteratively denoises $z_{t-1}$ from $z_t$ by first computing predicted clean latent $\hat{z}_0$ with $z_t$ and class label $y$:
\begin{align}
\hat{z}_0=\frac{\boldsymbol{z}_t-\sqrt{1-\alpha_t} \epsilon_\theta\left(\boldsymbol{z}_t,t,y\right)}{\sqrt{\alpha_t}} \label{eq:z0}
\end{align}
and then sampling $z_{t-1}$ based on $\hat{z}_0$: 
\begin{align}
\boldsymbol{z}_{t-1}&=\sqrt{\alpha_{t-1}}\cdot\hat{z}_0 +\sqrt{1-\alpha_{t-1}-\sigma_t^2} \cdot \epsilon_\theta\left(\boldsymbol{z}_t,t,y\right)+{\sigma_t \epsilon_t} \label{eq:DDIM}
\end{align}
where $\epsilon_t\sim\mathcal{N}(0,\mathrm{I})$ is standard Gaussian noise independent of $x_t$ and $\sigma_t$ controls the stochasticity of the update step.
\begin{algorithm}[h]
    \caption{Adversary-guided Curriculum Sampling}
    \label{alg:cag}
    \renewcommand{\algorithmicrequire}{\textbf{Input:}}
    \renewcommand{\algorithmicensure}{\textbf{Output:}}
    
    \begin{algorithmic}[1]
        \REQUIRE diffusion model $\epsilon_\theta$, number of curricula $N_c$, list of number of samples for each curriculum $[n_i]^{N_c}_{i=0}$, diffusion sampling steps $T$, vae decoder $D$
        \ENSURE distilled dataset $S$
        \FOR{$i=1$ to $N_c$}
            \STATE Train student model $\phi_i$ on $\bigcup_{k=0}^{i-1} S_k$ 
            \FOR{$j=0$ to $n_i$}
                \STATE Sample initial noise $z_T\sim\mathcal{N}(0,\mathrm{I})$
                \FOR{$t=T$ to $1$}
                    \STATE Denoise $z_{t-1}$ from $z_t$ by \eqref{eq:sample} with $\epsilon_\theta$ and $\phi_i$
                \ENDFOR
                \STATE $x_j=D(z_0)$
            \ENDFOR
            \STATE $S_i=\bigcup^{n_i}_{j=0} x_j$
        \ENDFOR
        \STATE $S=\bigcup_{i=0}^{N_c} S_i$
        
        \RETURN distilled dataset $S$
    \end{algorithmic}
\end{algorithm}

The total cost of diffusion-based dataset distillation involves fine-tuning the diffusion model on the target dataset and subsequently sampling from it. Consequently, this strategy circumvents the need for complex bi-level optimization procedure and the substantial increase in parameter volume that accompany high resolutions and image-per-class setting. Thus, it achieves an efficient distillation of large scale datasets\cite{Gu2023minimax}.

However, the inherent low temperature characteristic of diffusion sampling can result in similar typical patterns thus lack of diversity in distilled dataset. To address this issue, we introduce Adversary-guided Curriculum Sampling (ACS). Building upon the efficiency of the diffusion-based dataset distillation framework, ACS guides the diffusion sampling process toward generating sample $x_i$ that is complementary to the information already present in previous synthesized data. The overall framework of our method is shown in Figure \ref{fig:architecture}.

\subsection{Adversary-Guided Diffusion}

During the sampling stage of diffusion-based dataset distillation, given a dataset $\mathcal{S’}=\{x_j,y_j\}_{j=1}^{N_s'}$ ($N_s'<N_s$) which has already been generated, we are now sampling a new image $x_i$. Our objective is to ensure that $x_i$ introduces minimal informational overlap with the existing synthetic dataset $\mathcal{S’}$. To accomplish this, we posit that samples causing misclassification by a discriminator trained on $\mathcal{S’}$ provide information that is complementary to $\mathcal{S’}$. Accordingly, we commence by training a discriminator $\phi$ on synthetic dataset $\mathcal{S’}$. 
% %with cross-entropy loss:
% \begin{equation}
%     \phi=\underset{\phi}{\arg \min } \mathcal{L}_{\mathrm{ce}}\left(f_\phi(x_j), y_j\right), \quad \{x_j,y_j\}\in \mathcal{S’} \label{eq:stu train}
% \end{equation}
The objective of maximizing the informational complementarity between the current sampling instance $x_i$ and the existing dataset $\mathcal{S’}$ can be represented as minimizing an adversarial loss:
\begin{equation}
    \mathcal{L}_{adv}(x_i,y_i)=-\mathcal{L}_{ce}(f_{\phi}(x_i), y_i) \label{eq:advloss}
\end{equation}

\begin{figure}[h]
    \centering
    \includegraphics[width=9cm]{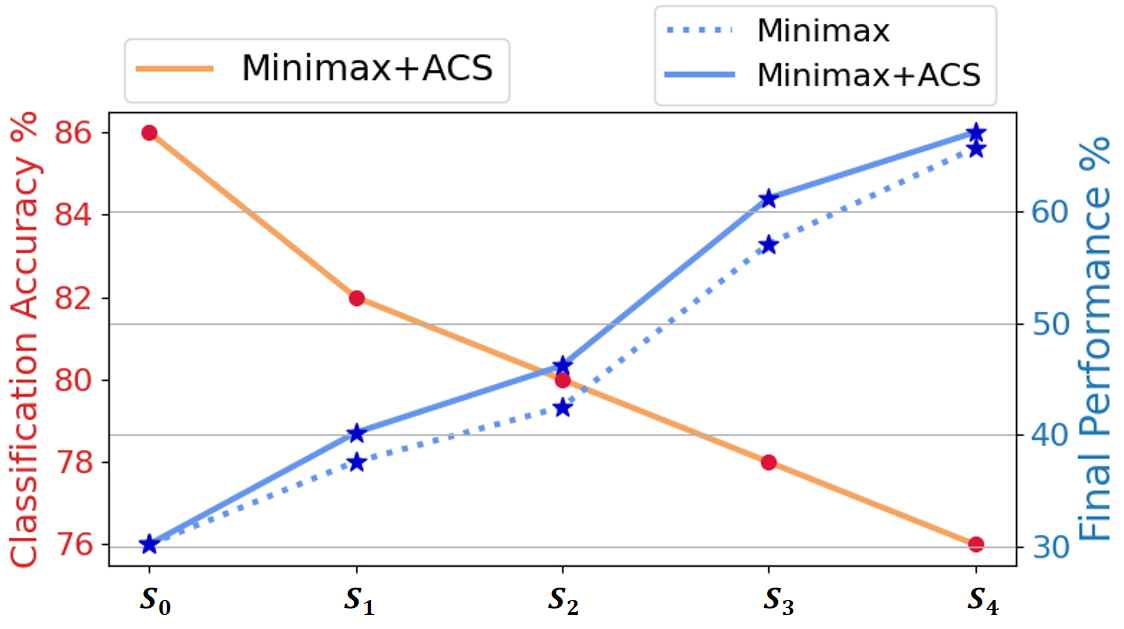}
    \caption{As the curricula advances, samples become increasingly hard to classify for a pre-trained classification model. Such simple to complex generation paradigm ensures a systematic and efficient coverage of target dataset informational spectrum thus lead to higher final test performance.}
    \label{fig:valacc}
\end{figure}

To introduce above objective into the generation process of $x_i$ , we employ guided diffusion \cite{adm, uniguidance, freedom} to guide the sampling trajectory of $x_i$ toward regions that minimize the adversarial loss within the target data manifold. Building upon the single-step sampling formula of DDIM in \eqref{eq:DDIM}, denoted as $z'_{t-1}=s(z_t, t, y, \boldsymbol{\epsilon}_\theta)$, adversary-guided diffusion sampling at each step can be formulated as:
\begin{equation}
    \boldsymbol{z}_{t-1}=s\left(\boldsymbol{z}_t, t, y,  \boldsymbol{\epsilon}_\theta\right)-s(t) \nabla_{\boldsymbol{z}_t} \mathcal{L}_{adv}\left({D(\hat{z}_0)},y_i\right)  \label{eq:sample}
\end{equation}

where $s(t)=g \cdot \frac{\sqrt{1-\alpha_t}\cdot||\epsilon_\theta\left(\boldsymbol{z}_t,t,c\right)||}{||\nabla_{\boldsymbol{z}_t} \mathcal{L}_{adv}\left({D(\hat{z}_0)},y_i\right)||}$ controls the guidance strength for each sampling step, with hyper-parameter $g$. $\hat{z}_0$ is the predicted final clean latent code of current $z_t$ with DDIM. Since diffusion sampling is carried in latent space, $D$ maps latent code $\hat{z}_0$ back to pixel domain which can be processed by discriminator $\phi$. Generally, adversarial guidance brings the generated image $x_i$ complementary to $\mathcal{S’}$ while ensures $x_i$ remains within the manifold of the target dataset $\mathcal{T}$.  

\subsection{Curriculum Diffusion Sampling}
% why call it curriculum?
With adversary-guided diffusion, ACS divides the distilled dataset into multiple curricula, eventually containing samples from simple to complex. Split distilled dataset can be represented as $S=\bigcup_{i=0}^{n} S_i$, where $n$ is the total number of curricula. 

ACS aggregate several other than single distilled sample as a curriculum because adding a single image $x_i$ to dataset $S'$ has limited impact on the performance of discriminator and time cost of training a discriminator is significant compared to sampling an image. Specifically, images in first curriculum $S_0$ are sampled without adversarial guidance. For generating samples in curriculum $S_i (i>0)$, we first train a discriminator $\phi_{i-1}$ on $S'=\bigcup_{j=0}^{i-1} S_j$ with cross entropy loss. Then, we employ $\phi_i$ for adversary-guided diffusion following Equation \eqref{eq:sample}. The entire process is outlined in the form of pseudo code as Algorithm \ref{alg:cag}.

\begin{figure}
    \centering
    \includegraphics[width=8.5cm]{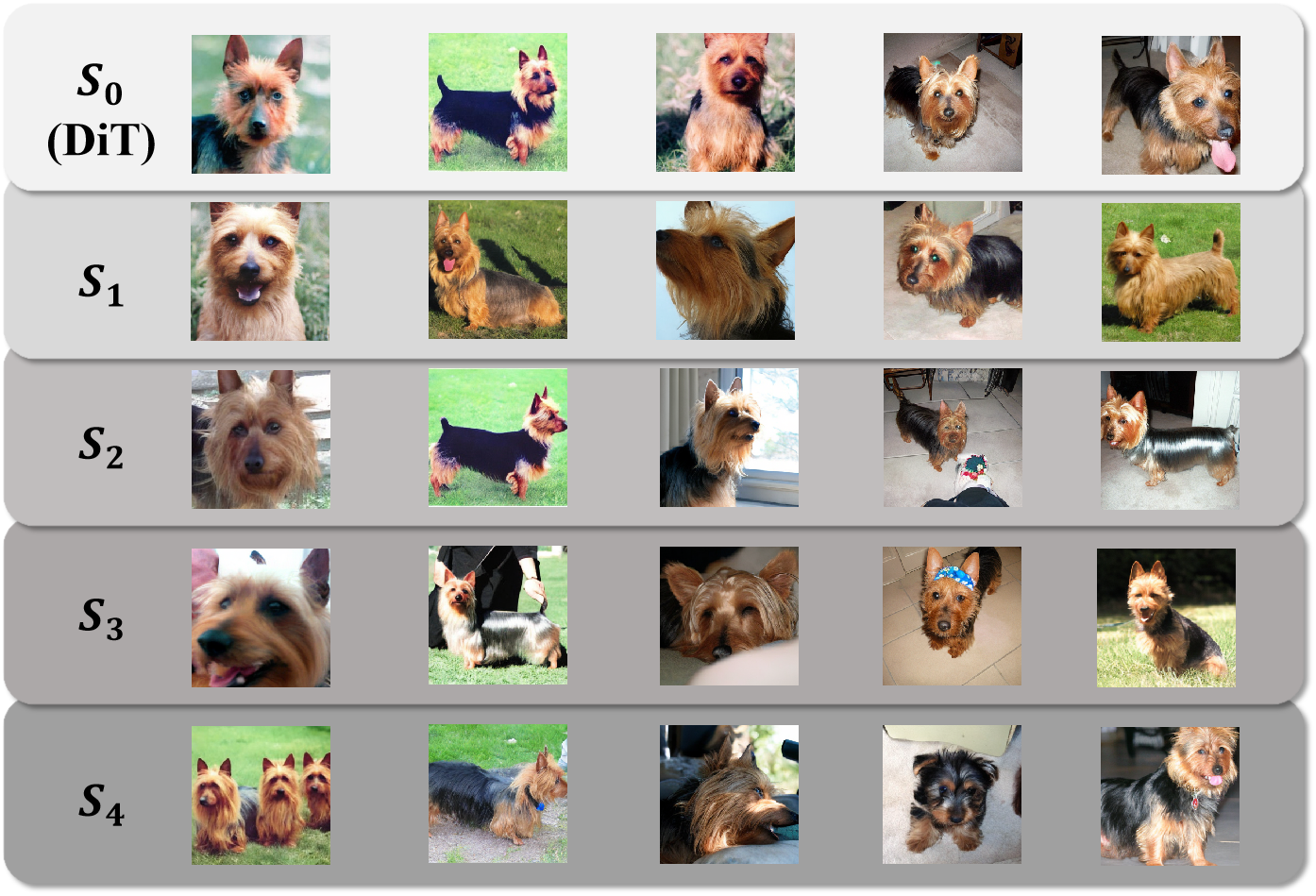}
    \caption{Visualization of Yorkshire Terrier images from different curricula. Later-stage samples exhibit rare characteristic like complex backgrounds, groupings of class subjects rather than single individuals,  localized or less distinctive close-ups, and rare subcategories such as juvenile instances.}
    \label{fig:vis}
\end{figure}

\begin{figure}
    \centering
    \includegraphics[width=8.5cm]{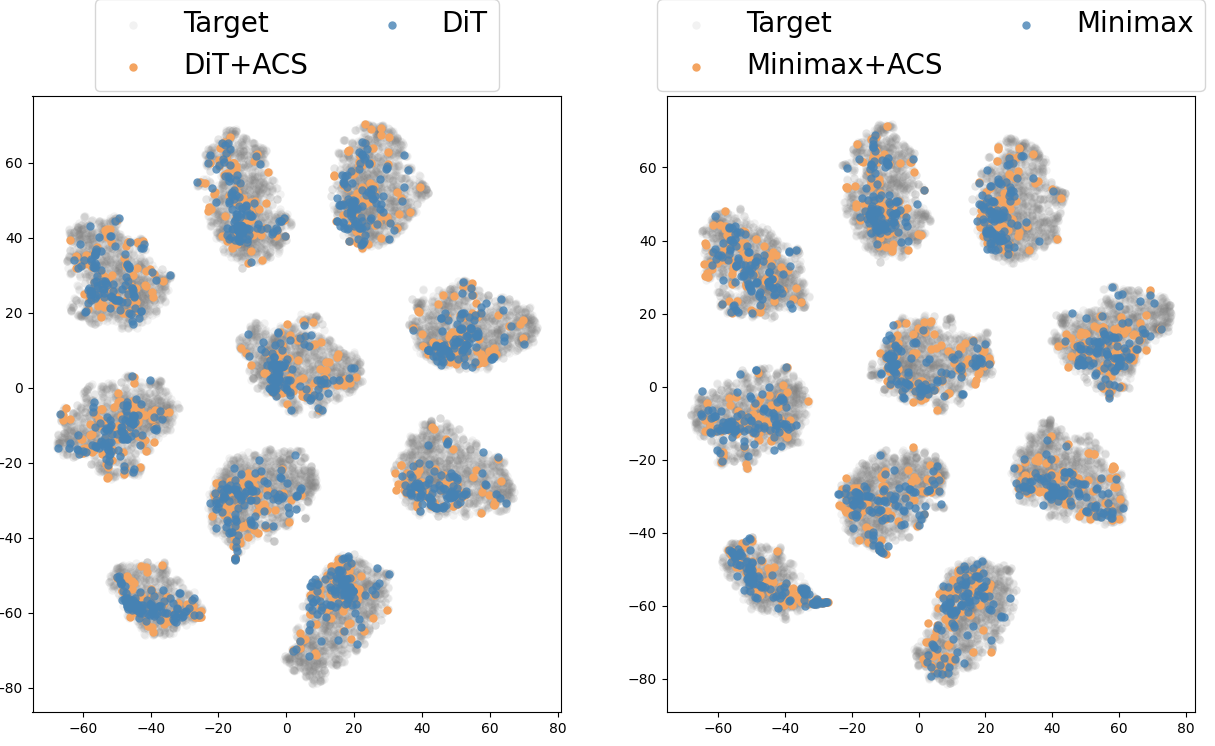}    \caption{t-SNE\cite{tsne} visualization of distilled dataset feature distribution generated with and without our ACS. With the help of ACS, distilled data better cover target feature distribution.}
    \label{fig:tsne}
\end{figure}

Recent works\cite{Selmatch, DATM, cudd, chen2025influenceguided} highlight the importance of increasing the complexity and diversity of the synthetic dataset as the number of IPC grows, ensuring an appropriately calibrated level of difficulty. Our proposed method facilitates this progression by establishing curriculum that transitions from simple to more challenging samples as the number of curricula expands. Initially, since diffusion models exhibit a tendency to overfit to the dense regions of the target data manifold, generated images focus on common and simple patterns\cite{Gu2023minimax, genlowdensity}. As the curriculum evolves, the performance of discriminator enhances, enabling it to accurately classify these easy samples. This leads adversary-guided diffusion to generate progressively rarer and more complex patterns. As illustrated in Figure \ref{fig:valacc}, the classification accuracy of a pre-trained classifier on distilled samples progressively diminishes as the curriculum advances, indicating increased complexity. Figure \ref{fig:vis} visualizes images of class Yorkshire Terrier from different curricula for each row. As the curriculum progresses, it becomes evident that samples in later curriculum exhibit characteristics of hard examples. These features can be summarized as follows: more complex backgrounds, groupings of class subjects rather than single individuals, extremely localized or less distinctive close-ups, and rare subcategories such as juvenile instances. With the help of ACS, distilled dataset better covers the distribution of target dataset, which is shown in Figure \ref{fig:tsne}.

\section{EXPERIMENTS}

\subsection{Experimental Setup}

\noindent \textbf{Datasets.}
Considering that the primary emphasis of our methodology lies in the distillation of high-resolution, large-scale datasets, we conduct distillation experiments on the ImageNet-1k dataset, alongside its subset ImageWoof\cite{ImageNet}. ImageWoof serves as a more rigorous benchmark for assessing the robustness and effectiveness of the distillation algorithm, given the subtle distinctions between its constituent classes\cite{Gu2023minimax}.

\noindent \textbf{Evaluation protocol and baselines.}
For fair comparison, we strictly follow the evaluation protocol of Minimax\cite{Gu2023minimax}. Specifically, we train multiple types of models on each distilled dataset with different image-per-class (IPC) setting. We report top-1 accuracy of each trained model on original validation set. Higher accuracy indicates superior performance of the dataset distillation algorithm. Each experiment is conducted 3 times, with the mean value and standard variance reported. Due to the scalability limitation of different dataset distillation methods, we compare with current state-of-the-art methods DM\cite{DM}, IDC-1\cite{IDC}, GLaD\cite{glad}, DiT\cite{DiT} and Minimax\cite{Gu2023minimax} on ImageWoof and DWA\cite{DWA}, CuDD\cite{cudd}, RDED\cite{Sun2023RDED}, DiT\cite{DiT} and Minimax \cite{Gu2023minimax} on ImageNet-1k. 

\noindent \textbf{Implement details.}
We leverage DiT\cite{DiT} and Minimax\cite{Gu2023minimax} as the backbone for diffusion-based dataset distillation, upon which we implement our Adversary-guided Curriculum Sampling. We set the diffusion denoising step number as 50. For curriculum diffusion sampling under IPC 100 setting, we simply divide the distilled dataset into 5 curricula and the number of samples for each curriculum as 5, 5, 10, 30, and 50, respectively. Then, the combination of the first two curricula constitutes the IPC 10 distilled dataset, the aggregation of the first three curricula forms the IPC 20 distilled dataset, and the union of the first four curricula comprises the IPC 50 distilled dataset. The model architecture of discriminator in Resnet10\cite{resnet} with average pooling.
\begin{table*}[h]
\centering
\setlength{\tabcolsep}{6pt}
\renewcommand{\arraystretch}{1.2}
\caption{Performance comparison with diffusion-based dataset distillation and other state-of-the-art methods on ImageWoof. The missing results are due to out-of-memory. The best results are marked as \textbf{bold}.}
\begin{tabular}{cc|cccccccccc}
\toprule
IPC(Ratio)           & Test Model  & Random          & DM\cite{DM}              & IDC-1\cite{IDC}           & GLaD\cite{glad}            & DiT\cite{DiT}             & DiT+ACS         & Minimax\cite{Gu2023minimax}         & Minimax+ACS        \\ \midrule
\multirow{3}{*}{10(0.8\%)}  & ConvNet-6   & $24.3_{\pm1.1}$  & $26.9_{\pm1.2}$ & $33.3_{\pm1.1}$ & $33.8_{\pm0.9}$ & $34.2_{\pm1.1}$ & $34.5_{\pm1.2}$ & $37.0_{\pm1.0}$ & $\mathbf{37.1_{\pm0.8}}$    \\
                            & ResNetAP-10 & $29.4_{\pm0.8}$  & $40.3_{\pm1.2}$ & $39.1_{\pm0.5}$ & $32.9_{\pm0.9}$ & $34.7_{\pm0.5}$ & $37.9_{\pm0.8}$ & $39.2_{\pm1.3}$ & $\mathbf{39.2_{\pm0.3}}$    \\
                            & ResNet-18   & $27.7_{\pm0.9}$  & $33.4_{\pm0.7}$ & $37.3_{\pm0.2}$ & $31.7_{\pm0.8}$ & $34.7_{\pm0.4}$ & $35.9_{\pm0.3}$ & $37.6_{\pm0.9}$ & $\mathbf{39.5_{\pm0.7}}$    \\ \midrule
\multirow{3}{*}{20(1.6\%)}  & ConvNet-6   & $29.1_{\pm0.7}$  & $29.9_{\pm1.0}$ & $35.5_{\pm0.8}$ &       -         & $36.1_{\pm0.8}$ & $38.1_{\pm0.7}$ & $37.6_{\pm0.2}$ & $\mathbf{39.3_{\pm0.1}}$    \\
                            & ResNetAP-10 & $32.7_{\pm0.4}$  & $35.2_{\pm0.6}$ & $43.4_{\pm0.3}$ &       -         & $41.1_{\pm0.8}$ & $44.9_{\pm1.1}$ & $45.8_{\pm0.5}$ & $\mathbf{45.9_{\pm0.5}}$    \\
                            & ResNet-18   & $24.3_{\pm1.1}$  & $29.8_{\pm1.7}$ & $38.6_{\pm0.2}$ &       -         & $40.5_{\pm0.5}$ & $41.9_{\pm0.5}$ & $42.5_{\pm0.6}$ & $\mathbf{46.2_{\pm0.2}}$    \\ \midrule
\multirow{3}{*}{50(3.8\%)}  & ConvNet-6   & $41.3_{\pm0.6}$  & $44.4_{\pm1.0}$ & $43.9_{\pm1.2}$ &       -         & $46.5_{\pm0.8}$ & $49.9_{\pm0.5}$ & $53.9_{\pm0.6}$ & $\mathbf{56.1_{\pm0.6}}$    \\
                            & ResNetAP-10 & $47.2_{\pm1.3}$  & $47.1_{\pm1.1}$ & $48.3_{\pm1.0}$ &       -         & $49.3_{\pm0.2}$ & $55.0_{\pm0.9}$ & $56.3_{\pm1.0}$ & $\mathbf{59.7_{\pm1.3}}$    \\
                             & ResNet-18  & $47.9_{\pm1.8}$  & $46.2_{\pm0.6}$ & $48.3_{\pm0.8}$ &       -         & $50.1_{\pm0.5}$ & $55.0_{\pm0.7}$ & $57.1_{\pm0.6}$ & $\mathbf{61.2_{\pm0.6}}$    \\ \midrule
\multirow{3}{*}{100(7.7\%)} & ConvNet-6   & $52.2_{\pm0.4}$  & $55.0_{\pm1.3}$ & $53.2_{\pm0.9}$ &       -         & $53.4_{\pm0.3}$ & $56.1_{\pm0.6}$ & $61.1_{\pm0.7}$ & $\mathbf{64.3_{\pm0.8}}$    \\
                            & ResNetAP-10 & $59.4_{\pm1.0}$  & $56.4_{\pm0.8}$ & $56.1_{\pm0.9}$ &       -         & $58.3_{\pm0.8}$ & $61.7_{\pm1.0}$ & $64.5_{\pm0.2}$ & $\mathbf{65.2_{\pm0.5}}$    \\
                            & ResNet-18   & $61.5_{\pm1.3}$  & $60.2_{\pm1.0}$ & $58.3_{\pm1.2}$ &       -         & $58.9_{\pm1.3}$ & $61.7_{\pm1.3}$ & $65.7_{\pm0.4}$ & $\mathbf{67.1_{\pm0.8}}$    \\ 
\bottomrule
\end{tabular}
\label{imagewoof}
\end{table*}
\begin{table*}[h]
\caption{Performance comparison on ImageNet-1K. The test model is Resnet18\cite{resnet}.}
\label{Imagenet1k}
\renewcommand{\arraystretch}{1.0}
\setlength{\tabcolsep}{6pt}
\centering
\begin{tabular}{c|ccccccccc}
\toprule
   IPC    &DWA\cite{DWA}   &  CuDD\cite{cudd}   & RDED\cite{Sun2023RDED}             & DiT\cite{DiT}             & DiT+ACS         & Minimax\cite{Gu2023minimax}         & Minimax+ACS \\ \midrule
10        &$37.9_{\pm0.2}$ &  $39.0_{\pm0.4}$   &$42.0_{\pm0.1}$ & $39.6_{\pm0.4}$ & $ 43.7_{\pm0.5} $ & $44.3_{\pm0.5}$ & $\mathbf{46.4_{\pm0.2}}$    \\ \midrule
50  &$55.2_{\pm0.4}$ &  $57.4_{\pm0.2}$   &$56.5_{\pm0.1}$ & $52.9_{\pm0.6}$ & $ 59.9_{\pm0.1} $ & $58.6_{\pm0.3}$ & $\mathbf{60.7_{\pm0.1}}$    \\
\bottomrule
\end{tabular}
\end{table*}
\subsection{Main Results}
 In our work, we introduce Adversary-guided Curriculum Sampling (ACS) into the sampling process of diffusion-based methods DiT\cite{DiT} and Minimax\cite{Gu2023minimax}, denoted as DiT+ACS and Minimax+ACS. The evaluation results on challenging dataset ImageWoof are shown in Table \ref{imagewoof}.

\noindent \textbf{Comparision with bi-level optimization-based methods.}
Bi-level optimization-based dataset distillation IDC-1\cite{IDC} and DM\cite{DM} exhibit highly efficient compression performance at lower IPC levels such as IPC 10 and IPC 20. However, considering that the marginal benefits of increasing data volume on model training performance diminish as the dataset size grows, random sampling serves as a strong baseline under high IPC setting. As the IPC increases to 50 and 100, the performance of DM and IDC-1 tends to align with or even fall short of that achieved by Random Selection. GLaD\cite{glad} is unable to scale up to the IPC 20 setting due to its substantial computational demands. In contrast, the diffusion-based method Minimax\cite{Gu2023minimax} with our ACS, not only demonstrates significant performance at smaller IPC levels but also maintains superior compression effectiveness at higher IPC levels. 

\noindent \textbf{Comparision with diffusion-based methods.}
ACS significantly enhances the performance across various compression ratio settings. At lower IPC levels (IPC 10 and IPC 20), our method achieves an average improvement of 2\% compared to DiT\cite{DiT} and 1.2\% relative to Minimax\cite{Gu2023minimax}. Note that at these lower IPC levels, the low-temperature characteristics inherent in diffusion sampling have a less pronounced effect, and the special designed loss introduced by Minimax during training partially mitigates this issue. As the IPC increases, the low-temperature characteristics become more prominent, and the performance gains from our method become increasingly significant. In high IPC scenarios, our method manages to achieve an average improvement of 3.8\% over DiT and 2.1\% over Minimax. This underscores the ability of our method to generate information-rich distilled dataset with the same volume of data, highlighting its efficiency. Furthermore, consistent improvements across all types of test model demonstrate the robustness of our approach. Experiment results on ImageNet-1k is shown in Table \ref{Imagenet1k}. Our method achieves an average improvement of 5.6\% over DiT and 2.1\% over Minimax, demonstrating the scalability of our approach.
\begin{table}[]
\caption{Hyper-parameter analysis on number of curricula. The result are obtained with Resnet18\cite{resnet} on ImageWoof.}
\centering
\renewcommand{\arraystretch}{1.0}
\begin{tabular}{@{}c|cccc@{}}
\toprule
Number of Curricula & 1                & 2 & 3 & 4 \\ \midrule
DiT+ACS             & $50.1_{\pm0.5}$  & $52.7_{\pm0.6}$  & $53.0_{\pm0.6}$  & $\mathbf{55.0_{\pm0.7}}$ \\
Minimax+ACS         & $57.1_{\pm0.6}$  & $58.8_{\pm0.3}$  & $59.6_{\pm0.8}$  & $\mathbf{61.2_{\pm0.6}}$ \\ 
\bottomrule
\end{tabular}
\label{tab:curricula num}
\end{table}
\subsection{Analysis}

\noindent \textbf{Analysis on numbers of curricula.}
Table \ref{tab:curricula num} delineates the impact of varying numbers of curricula on the test performance when ACS is applied to DiT\cite{DiT} and Minimax\cite{Gu2023minimax} for IPC 50 distillation. A single curriculum scenario serves as the baseline, wherein ACS is not introduced. By bifurcating the generation of the synthetic dataset into two curricula and incorporating adversary-guided diffusion within the second curriculum, we observe respective performance enhancements of 2.6\% and 1.7\% for DiT and Minimax. Moreover, the progressive augmentation in the number of curricula yields a corresponding incremental improvement in performance. This trend underscores the efficacy of integrating adversary-guided diffusion into the distilled dataset sampling process through a curricular approach. The granularity of the curriculum configuration can be meticulously tuned to achieve an optimal balance between performance and efficiency.

\begin{figure}
    \centering
    \includegraphics[width=7cm]{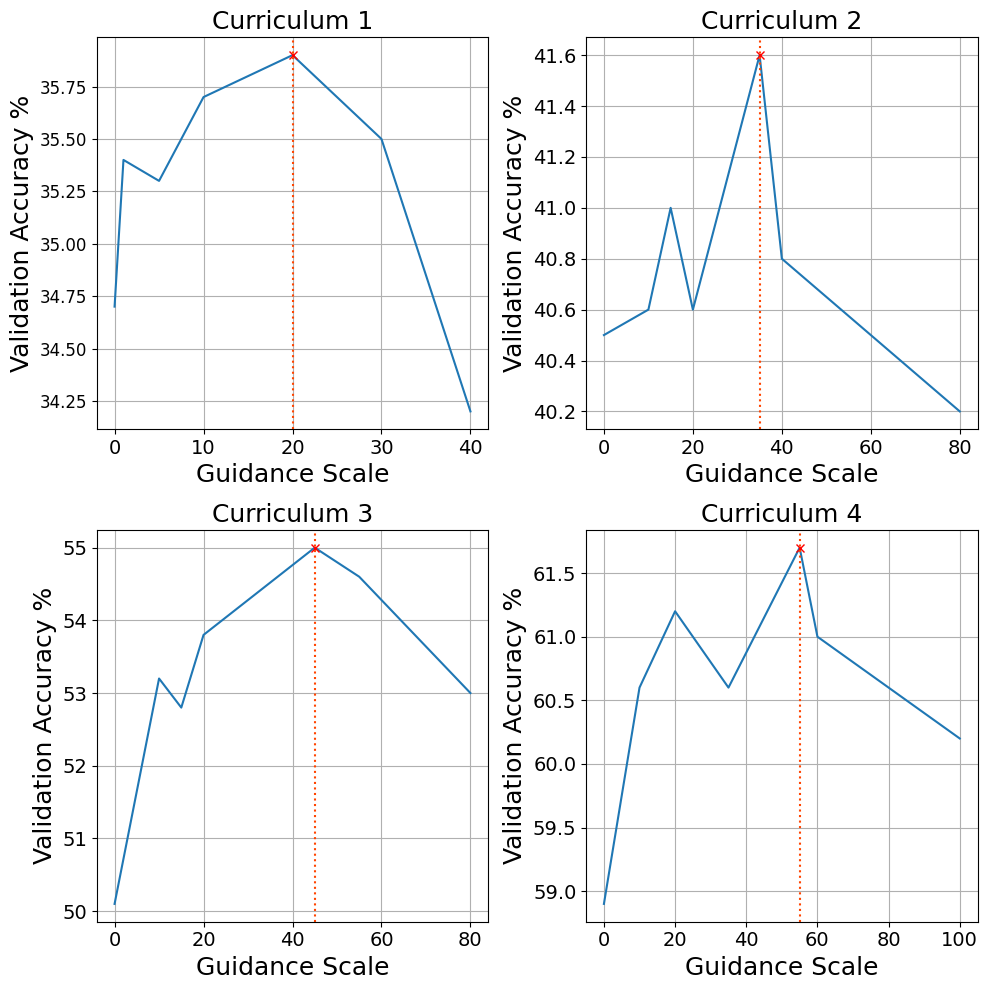}
    \caption{Hyper-parameter analysis on guidance scale $g$ in different curriculum. The result are obtained with Resnet18\cite{resnet} on ImageWoof by DiT+ACS.}
    \label{fig:guidance scale analysis}
\end{figure}

\noindent \textbf{Analysis on adversarial guidance scale.}
To substantiate the efficacy of adversary-guided diffusion and its influence on the sampling process, Figure \ref{fig:guidance scale analysis} illustrates the variations in test performance with different guidance scales $g$ in \eqref{eq:sample} of DiT+ACS across four curricula. A guidance scale of zero represents the baseline without ACS. For all curricula, an increase in the guidance scale from zero generally corresponds with improved performance. This observation corroborates the enhancement that adversary-guided diffusion can provide to diffusion-based methods during the sampling phase. However, when the intensity of adversarial guidance becomes excessive, it may introduce samples that are overly complex or rare for the current curriculum, even deviating from the target data manifold. Such excess undermine the inherent advantages of diffusion-based methods, leading to a decline in performance, which may eventually fall below baseline levels.

\section{CONCLUSIONS}
In this work, we introduce Adversary-guided Curriculum Sampling for diffusion-based dataset distillation. ACS divides the synthesis of distilled dataset into multiple curricula and guides the diffusion sampling process of individual image in each curriculum with an adversarial loss towards generating patterns that are increasingly challenging for a discriminator trained on preceding curricula. It achieves state of the art on large scale and high resolution ImageNet-1k and its challenging subset.

\bibliographystyle{IEEEbib}
\bibliography{icme2025references}

\end{document}